\documentclass[conference]{IEEEtran}
\IEEEoverridecommandlockouts
\usepackage{cite}
\usepackage{amsmath,amssymb,amsfonts}
\usepackage{algorithmic}
\usepackage{graphicx}
\usepackage{textcomp}
\usepackage{xcolor}
\usepackage{hyperref}
\usepackage{booktabs}
\usepackage{multirow}
\usepackage{subcaption}
\usepackage{amsmath}
\usepackage{bm}
\usepackage{svg}
\usepackage{tabularx}
\usepackage[linesnumbered,ruled]{algorithm2e}
\usepackage{siunitx}

\def\BibTeX{{\rm B\kern-.05em{\sc i\kern-.025em b}\kern-.08em
    T\kern-.1667em\lower.7ex\hbox{E}\kern-.125emX}}
\begin{document}

\title{Affective Priming Score: A Data-Driven Method to Detect Priming in Sequential Datasets \\
\thanks{This work was supported by the Wallenberg AI, Autonomous Systems and Software Program (WASP) funded by the Knut and Alice Wallenberg Foundation.}
}

\author{\IEEEauthorblockN{Eduardo Gutierrez Maestro}
\IEEEauthorblockA{\textit{Centre for Applied Autonomous} \\ \textit{Sensor Systems (AASS)} \\
\textit{Örebro University, Sweden}\\
\href{mailto:eduardo.gutierrez-maestro@oru.se}{eduardo.gutierrez-maestro@oru.se}}
\and

\IEEEauthorblockN{Hadi Banaee}
\IEEEauthorblockA{\textit{Centre for Applied Autonomous} \\ \textit{Sensor Systems (AASS)} \\
\textit{Örebro University, Sweden}\\
\href{mailto:hadi.banaee@oru.se}{hadi.banaee@oru.se}}
\and

\IEEEauthorblockN{Amy Loutfi}
\IEEEauthorblockA{\textit{Centre for Applied Autonomous} \\ \textit{Sensor Systems (AASS)} \\
\textit{Örebro University, Sweden}\\
\href{mailto:amy.loutfi@oru.se}{amy.loutfi@oru.se}}
}

\maketitle

\begin{abstract}
Affective priming exemplifies the challenge of ambiguity in affective computing. While the community has largely addressed this issue from a label-based perspective, identifying data points in the sequence affected by the priming effect, the impact of priming on data itself, particularly in physiological signals, remains underexplored. Data affected by priming can lead to misclassifications when used in learning models. This study proposes the Affective Priming Score (APS), a data-driven method to detect data points influenced by the priming effect. The APS assigns a score to each data point, quantifying the extent to which it is affected by priming. To validate this method, we apply it to the SEED and SEED-VII datasets, which contain sufficient transitions between emotional events to exhibit priming effects. We train models with the same configuration using both the original data and priming-free sequences. The misclassification rate is significantly reduced when using priming-free sequences compared to the original data. This work contributes to the broader challenge of ambiguity by identifying and mitigating priming effects at the data level, enhancing model robustness, and offering valuable insights for the design and collection of affective computing datasets.
\end{abstract}


\section{Introduction}

Ambiguity is an increasingly important research challenge in affective computing. Researchers from multiple disciplines tackle this issue from different perspectives. In psychology, ambiguity arises in various contexts, such as designing stimuli that effectively engage participants or defining and labeling emotional concepts consistently.
In computer science, ambiguity becomes a challenge in machine learning when labels are subjective and do not precisely represent the underlying data, as is often the case when labeling data with emotions.

Affective priming is a concrete case of ambiguity in affective computing, where the order of emotional events influences how subjects perceive and react to subsequent stimuli~\cite{bargh2000mind}. Prior studies have examined this effect primarily from a labeling perspective. For instance, authors demonstrated in~\cite{shen2019unintentional} that previously seen images biased annotators' judgments-neutral facial expressions were more likely to be labeled as sad if preceded by a sequence of sad expressions. Similarly, in~\cite{martinez2023analyzing}, authors analyzed how inter-annotator disagreement led to labels influenced by priming, ultimately affecting model performance.

Although the community has explored the effects of priming mainly from the label perspective, i.e., how priming manifests in labeled data, this effect also alters the underlying data itself. Some data modalities, like physiological signals, are highly susceptible to the priming effect. For example, in~\cite{maestro2023stress}, researchers observed statistically significant differences in model performance when trained on datasets in which a stressful task was induced before an amusement task, suggesting that priming is present in the data. However, their analysis was limited to data sequences containing only a single transition between emotional states, which is insufficient to generalize the presence of priming. Thus, the study of data-centric priming remains fairly unexplored.

\begin{figure}
    \centering
    \includegraphics[width=\linewidth]{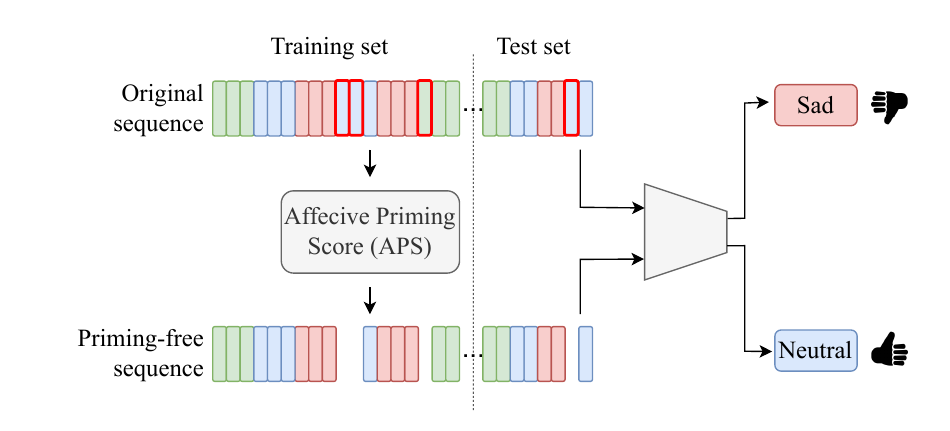}
    \caption{Affective priming can be observed in sequential datasets where a set of emotional categories is induced (happy: green, neutral: blue, sad: red). Models trained on data points affected by the priming effect (highlighted in red in the top sequence) tend to misclassify emotions.}
    \label{fig:graph-abstract}
\end{figure}

Our work addresses this gap by introducing the Affective Priming Score (APS), a method designed to identify data points affected by priming in sequential datasets. The APS assigns a score to each data point, quantifying the extent to which it has been influenced by the emotional event preceding it. We hypothesize that data points affected by priming are more prone to misclassification, particularly into the class of the preceding event. Figure~\ref{fig:graph-abstract} illustrates such a hypothesis. To validate this hypothesis, we introduce the Priming Error (PE) metric, which measures the proportion of data points misclassified as the previous event’s class. Since priming labels are not explicitly available in datasets, PE serves as an indirect validation tool.

Our experimental results show that training models on datasets where high-APS data points have been removed leads to a significant reduction in the PE metric, confirming the effectiveness of APS in detecting priming-affected data points. Additionally, APS provides a practical tool for dataset design, allowing researchers to visualize the extent to which data points might be influenced by the sequentiality of induced stimuli.

This study contributes to the broader challenge of ambiguity in affective computing by addressing the data-driven impact of affective priming. By identifying and mitigating priming effects at the data level, our approach enhances model robustness and provides insights for the design and collection of affective computing datasets.

\section{Related work}

% Related work on sequential models and priming. The goal is to highlight

\subsection{Priming in Affective Computing}

In affective computing, the priming effect has been explicitly explored in several works. 
In~\cite{shen2019unintentional}, the authors studied how priming affects biases in labeling facial expression images when presented in a sequence.
They investigated the impact of sequence order on labeling facial expressions and found a positive correlation: Participants were more likely to label a neutral image with the emotion corresponding to the preceding positive or negative images in the sequence.
The authors in~\cite{martinez2023analyzing} followed a similar path by proposing a metric to quantify the affective priming present in the annotation of the data in the speech data. 
Additionally, the authors used this metric to create subsets of different levels of priming and found better performance of models trained on less biased annotations. 
In machine learning, affective priming is associated with data samples being labeled with noisy or ambiguous information, which can hinder the performance of the model~\cite{martinez2023analyzing}.

The sequential order of events can also influence the data that is generated. For example, the physiological reaction to a stressful event may have so called lingering effects that influence the data collected during subsequent events.
In~\cite{maestro2023stress}, Maestro \textit{et al.} investigated this impact and found significant statistical differences in model performance between two groups.
In the first group, amusement stimuli appeared first in the sequence, whereas in the second group, stressful stimuli came first.
Performance in the second group dropped significantly compared to the first, supporting the hypothesis that the order of stimuli influences the data—i.e., affective priming occurs from a data perspective as well.
However, the data sequences used in~\cite{maestro2023stress} were short and contained only one transition between different emotional states, making it difficult to generalize the extent of the priming effect.

\subsection{Sequential models in Affective Computing}

The physiological response to an emotional stimulus generates data that contains temporal information.
Sequential models are designed to capture temporal relationships between data points in a sequence and extract meaningful patterns from time-series data.
In the literature, Long Short-Term Memory (LSTM)~\cite{hochreiter1996lstm} networks and transformer~\cite{vaswani2017attention} models are the most widely used architectures for learning tasks involving time-dependent data, with the latter representing the state-of-the-art approach.

In affective computing, these models are used to improve the prediction accuracy in different benchmarks. However, in the literature, there are studies that make use of such model architectures, where the temporality between samples in the sequence is not correctly considered. For example, in~\cite{du2020efficient}, the authors proposed an efficient LSTM network where the number of channels was used to define the temporality in the model's input data. Similar input data is proposed in~\cite{sakalle2021lstm}. In this case, the LSTM architecture was fed at each time step by one of the 20 features in each data point. 

There are other studies in the field where the model is fed by sequences of data, being able to learn patterns in the transitions between emotional states. However, the validation method chosen to evaluate the trained model may hide the difficulties present in affective sequential datasets. The stationary property of a signal is normally not the same throughout the transitions from one emotional state to another. In~\cite{meng2023eeg} and~\cite{ouyang2025daeegvit}, an LSTM and Transformer architecture was trained, respectively, where the input data was configured to capture temporal information in the sequence. However, the data were split randomly, making the learning task less challenging as the model would see data points for the same class where the stationarity was different.
Similarly, it happens in~\cite{zhang2024transformer}~\cite{song23Trans}, where the authors used the K-fold validation method.

%%%%%%%%

Compared with related work, this study studies the priming effect from a data-driven perspective, i.e., how priming impacts the data quality in sequential way. Additionally, importance is given to the validation method chosen, as the impact of the priming effect may be eluded when validation methods chosen that do not keep the temporality between the train and test sets. 

\section{Datasets} \label{sec:datasets}

This study employs the family of SEED~\cite{seed} datasets to validate the proposed method. 
Specifically, we focus on SEED~\cite{seed} and SEED-VII~\cite{seed7}. 
These datasets consist of sequences of audio-visual emotional stimuli, where the subjects' reactions to each stimulus are recorded. 
We select these benchmarks because they contain multiple transitions between emotional states. 
In contrast, other datasets in the field, such as DEAP~\cite{deap} and DREAMER~\cite{dreamer}, have insufficient emotional state transitions. 
For example, in the DREAMER~\cite{dreamer} dataset, nine emotions were induced, with each emotion elicited only twice in the sequence. 
The low number of transitions between emotional trials makes these datasets unsuitable for this study.

\begin{figure}
    \centering
    \includegraphics[width=\linewidth]{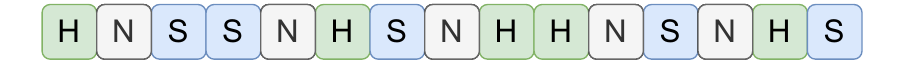}
    \caption{Sequence of emotional trials in SEED~(H:~Happy, N:~Neutral, S:~Sad).}
    \label{fig:seed}
\end{figure}

\textbf{SEED.}
The authors selected three emotional states to elicit in the subjects: happy~(H), neutral~(N), and sad~(S). The data collection protocol involves exposing each subject to a sequence of videos, with each video chosen to induce one of the three emotions. In each video, an Electroencephalogram~(EEG) signal is recorded. Each video exposure is referred to as a trial, and the full sequence consists of $15$ trials. The whole sequence of trials is defined as a session. This dataset encompasses a total of $15$ subjects.
Each subject participated in three sessions, which were recorded on different days. Fig.~\ref{fig:seed} illustrates the sequence of emotional trials across the three sessions. The variety of transitions between the three emotional states makes this dataset particularly suitable for studying the effects of priming from a data-driven perspective.

\begin{figure}[]
        \centering
           \subfloat[Session 1]{%
              \includegraphics[width=\linewidth]{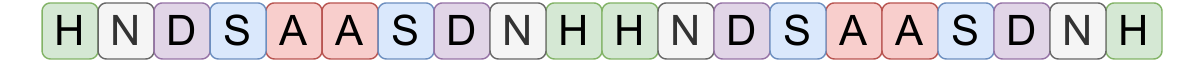}%
              % \label{fig:seed}%
           } 
           \vspace{0.2cm}
           \subfloat[Session 2]{%
              \includegraphics[width=\linewidth]{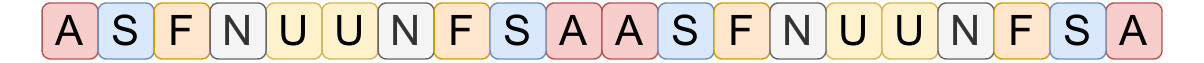}%
              % \label{fig:cm-seed-ldl}%
           }
            \vspace{0.2cm}
           \subfloat[Session 3]{%
              \includegraphics[width=\linewidth]{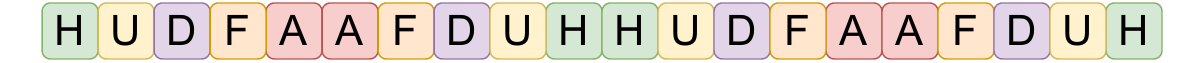}%
              % \label{fig:cm-seed-ldl}%
              % \label{fig:cm-seed}
           }
           \vspace{0.2cm} 
           \subfloat[Session 4]{%
              \includegraphics[width=\linewidth]{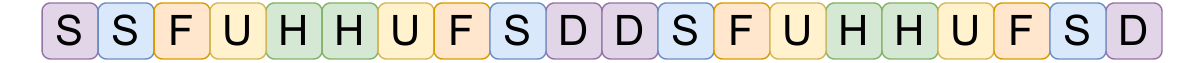}%
              % \label{fig:cm-seed-ldl}%
           }
           \caption{Sequence of emotional trials in SEED-VII in each session~(H:~Happy, N:~Neutral, S:~Sad, U:~Surprise, D:~Disgust, F:~Fear, A:~Anger).}
           \label{fig:seed7}           
\end{figure}

\textbf{SEED-VII.}
To further validate the proposed method, we select an additional benchmark: SEED-VII~\cite{seed7}, an extension of the SEED dataset.  
The key differences between these benchmarks lie in the total number of targeted emotions and the number of trials per session.  
SEED-VII includes seven emotions: happy~(H), surprise~(U), neutral~(N), disgust~(D), fear~(F), sad~(S), and anger~(A).  
These emotions are elicited across four sessions, with each session consisting of $20$ trials—that is, each subject is exposed to $20$ emotional video clips per session. This dataset in total encompasses $20$ subjects.  
To prevent abrupt shifts in emotional valence, the authors designed the sequence of video clips to ensure a smoother transition between emotional states~\cite{seed7}. Fig.~\ref{fig:seed7} illustrates the sequence of video clips in each session of the benchmark. According to the authors, the order of the clips was carefully designed to minimize abrupt emotional shifts, making this benchmark particularly challenging for validating the proposed method.  

For both benchmarks, we use the Differential Entropy~(DE) features provided by the authors.
A sliding window technique is used to extract features from each EEG segment at five frequencies bands (delta: \SIrange{1}{3}{\hertz}, theta: \SIrange{4}{7}{\hertz}, alpha: \SIrange{8}{13}{\hertz}, beta: \SIrange{14}{30}{\hertz}, gamma: \SIrange{31}{50}{\hertz}). These frequency bands have been reported to perform well for emotion classification~\cite{seed, seed7}. Hence, each sliding window yields a feature matrix in $\mathbb{R}^{62 \times 5}$, where $62$ is the number of EEG channels and $5$ corresponds to the DE values computed per channel across the five frequency bands. 
We use the vector obtained by flattening this matrix as input to the proposed method.

\section{Methodology}
\subsection{Problem Formulation} \label{sec:problem}

\begin{figure*}[ht]
    \centering
    \includegraphics[width=0.8\linewidth]{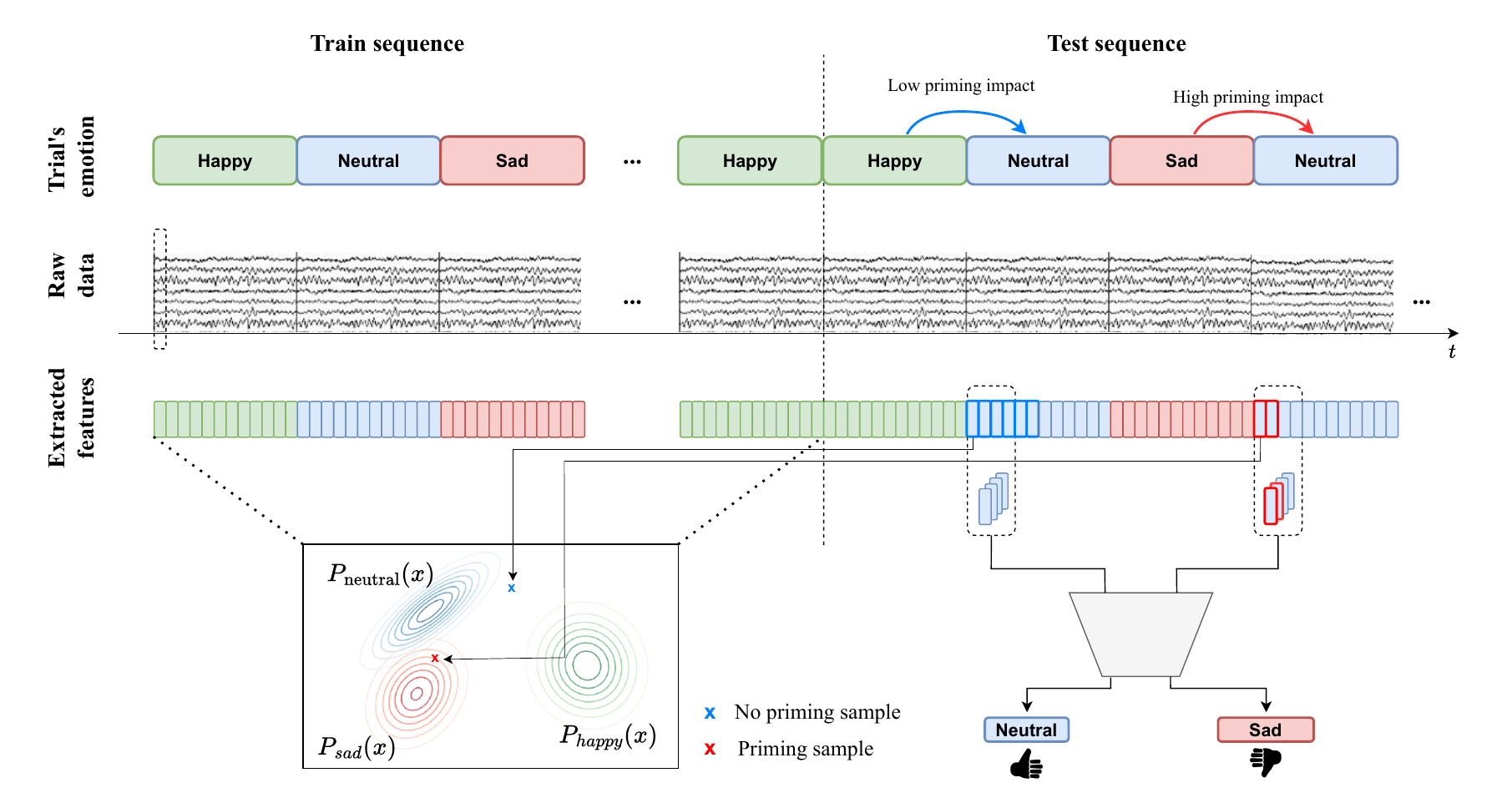}
    \caption{The data generated in response to an emotional trial may be influenced by the order in which these trials are presented. For example, a subject's reaction to a neutral trial may be less affected when preceded by a happy trial than when preceded by a sad trial. Consequently, data points affected by the priming effect (highlighted in red) tend to be closer to the posterior probability function of the trial that caused the priming, leading models to make erroneous predictions.}
    \label{fig:problem-formulation}
\end{figure*}

Consider a dataset that consists of a sequence of trials~$\{T_1, T_2, \dots, T_N\}$, where each trial is the exposure to an emotional stimulus, and $N$ is the total number of trials in the sequence.
The set $\mathcal{C} \in \{c_1, c_2, \dots, c_K \}$ represents all possible emotions induced across trials.  Let $y_{i,k}$ denote the emotion induced at trial $T_i$ in the sequence, corresponding to emotion $c_k$. 
Additionally, each trial $T_i$ is composed of a set of data points $T_i \in \{\mathbf{x}_{i,1}, \mathbf{x}_{i,2}, \dots, \mathbf{x}_{i, M}\}$, being $M$ the total number of data points in the trial. 
Each data point is a feature vector representing the subject’s response to the emotional stimulus and is labeled with the emotion associated with that trial, $y_{i,k}$.

Given the description above, suppose a dataset consists of $N$ trials.
Let this data be used to train a machine learning model to detect three different emotions: $ \{ \text{happy}, \text{neutral}, \text{sad} \}$. Let $\{P_{\text{happy}}(x),  P_{\text{neutral}}(x), P_{\text{sad}}(x) \}$ be the posterior probability functions learned by the model for each of the targeted emotions.

Due to the lingering nature of emotions, the order of the induced stimuli can impact the perception and reactions of subjects, a phenomenon known as affective priming. In machine learning, this translates to data points being projected closer to a posterior probability function that does not correspond to their actual label as learned by the model. This problem is illustrated in Fig.~\ref{fig:problem-formulation}. For example, in the trials of the test sequence, the \textit{neutral} trial appears twice: first following a \textit{happy} trial and later following a \textit{sad} trial. In the former case, data points in the \textit{neutral} trial suffer from low priming, whereas in the latter, data points in the neutral trial are strongly affected by the priming effect. As a result, the model is more likely to project priming-affected data points closer to the priming class posterior probability function, i.e., $P_{\text{sad}}(x)$ in this case, ultimately leading to incorrect predictions.
This study investigates how the sequential order of emotional trials influences data generation within a dataset.

\subsection{Affective Priming Score} \label{sec:score}

To address the problem of affective priming presented in section~\ref{sec:problem}, we introduce the Affective Priming Score (APS), a method that uses a data-driven approach to score data points affected by the priming effect in sequential datasets. 
Based on the formulation presented in section~\ref{sec:problem}, this method evaluates the impact of priming in each data point by measuring the distance to data points in past trials. A closer distance to data points in the preceding trial in the sequence than to data points of the same class in previous trials indicates an influence of the priming effect. 
The APS is calculated for each data point~($x_i$) within each trial $T_i$ using the following function:

\begin{equation} 
    \label{eq:aps}
    APS(x_i) = \frac{e^{-d(x_{i}, g_{prev})/\tau}}
    {\sum_{j \in \{class, prev, other\}} e^{-d(x_{i}, g_{j})/\tau}}
\end{equation}

% ==============================================================
% Hadi:

% \begin{equation} \label{eq:aps}
%     APS(x_i) = \frac{e^{-d(x_{i}, g_{prev})/\tau}}
%     {\sum_{j \in \{class, prev, other\}} e^{-d(x_{i}, g_{j})/\tau}}
% \end{equation}

% or:

% \begin{equation} \label{eq:aps}
%     APS(x_i) = \frac{e^{-d(x_{i}, g_{prev})/\tau}}
%     {\sum_{g_j \in \{g_{class}, g_{prev}, g_{other}\}} e^{-d(x_{i}, g_{j})/\tau}}
% \end{equation}

% ==============================================================

Eq.~\ref{eq:aps} sequentially computes the APS values. Specifically, for each trial in the sequence, the APS value is calculated for all data points within that trial. To achieve this, the following terms are needed: $\{ d(x_{i}, g_{class}), d(x_{i}, g_{prev}), d(x_{i}, g_{other}) \}$. These terms are computed using a distance function $d$; in this case, we use the Euclidean distance to a set of centroids. As mentioned, APS values are calculated for each trial $T_i$. Using the set $\{T_0, \dots, T_{i-1} \}$, we compute the centroids $\{ g_{\text{class}}, g_{\text{prev}}, g_{\text{other}} \}$. The centroid $g_{\text{class}}$ is computed using data points corresponding to the same class as $x_i$; $g_{\text{prev}}$ is computed using data points from the class of the previous trial to $x_i$; finally, $g_{\text{other}}$ is computed using data points from the other classes in past trials in the sequence. Eq.~\ref{eq:aps} applies the softmax function, aiming to output higher scores when the distances of the data points in the assessed trial are closer to the centroid of the class in the previous trial. In Eq.~\ref{eq:aps}, $\tau$ is a temperature parameter used to soften the softmax output. In this study, it is set to $0.1$. 

Additionally, this method makes the following assumptions: 1) The APS for the first $K$ trials is 0, where $K$ is the total number of induced emotions. This assumption is based on the fact that priming requires prior emotional exposure, which is absent at the beginning of the sequence.  2) If two consecutive trials induce the same emotional class, the APS values for the second trial are 0, as a trial of the same class cannot provoke priming by definition.

\section{Experimental results}

This section presents the results and benefits of applying the proposed method to data sequences affected by the priming effect. It is divided into two parts.

The first part examines the output of the APS method, presenting the averaged APS values computed across sequences. The second part validates the efficiency of the proposed method by comparing the performance of models trained on the original sequences with those trained on their priming-free versions.

\subsection{Affective Priming Score in sequential datasets}

We compute the affective priming score for each data point in the sequence using the method described in Section~\ref{sec:score}. The presented method assumes the data points of the initial trials to have zero APS value, thus, they are excluded from the illustrations presented in this section. In SEED and SEED-VII, the first three and five trials are excluded, respectively.  

\begin{figure}
    \centering
    \includegraphics[width=\linewidth]{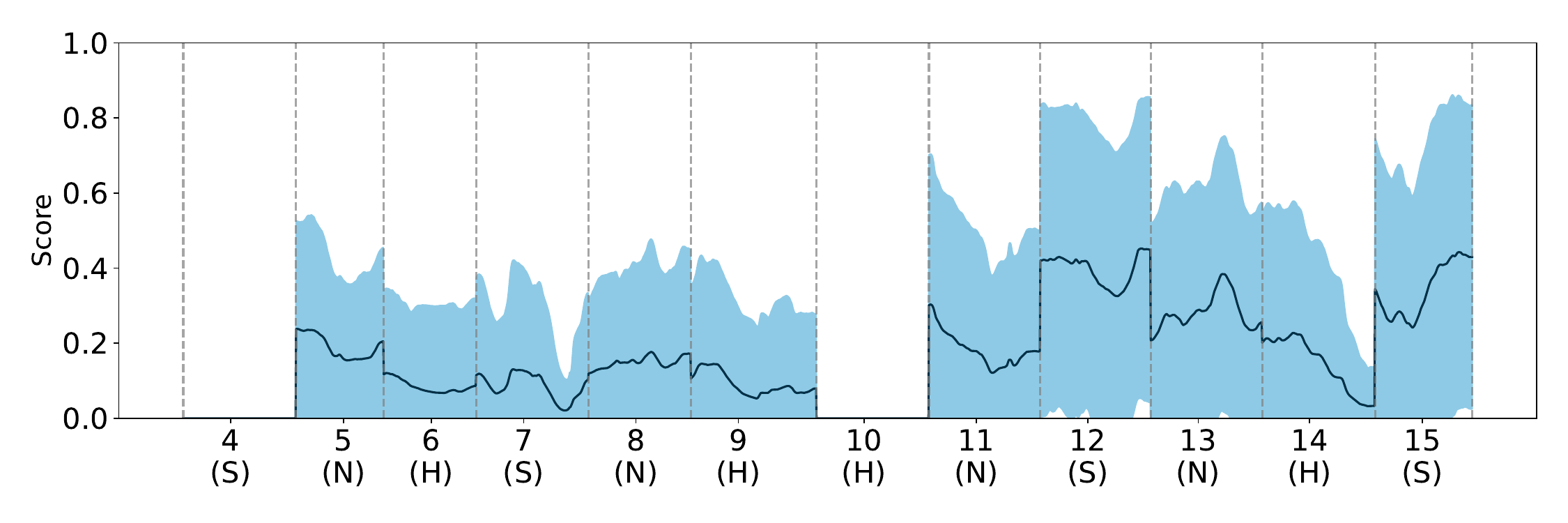}
    \caption{Averaged priming score for SEED dataset. Note that when two consecutive trials in the sequence share the same emotion, the affective priming score in the latter is assumed to be zero.}
    \label{fig:seed-average-score}
\end{figure}

\begin{figure}[t]
        \centering
           \subfloat[Session 1]{%
              \includegraphics[width=\linewidth]{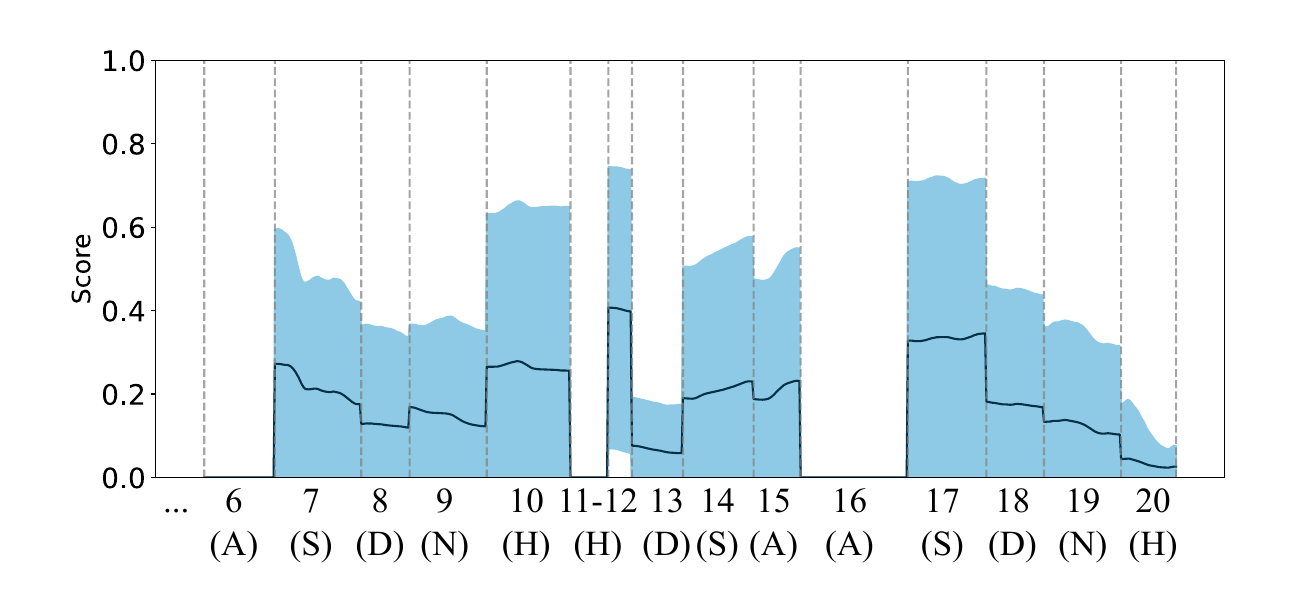}%
           } 
           
           \vspace{0.2cm}
           
           \subfloat[Session 2]{%
              \includegraphics[width=\linewidth]{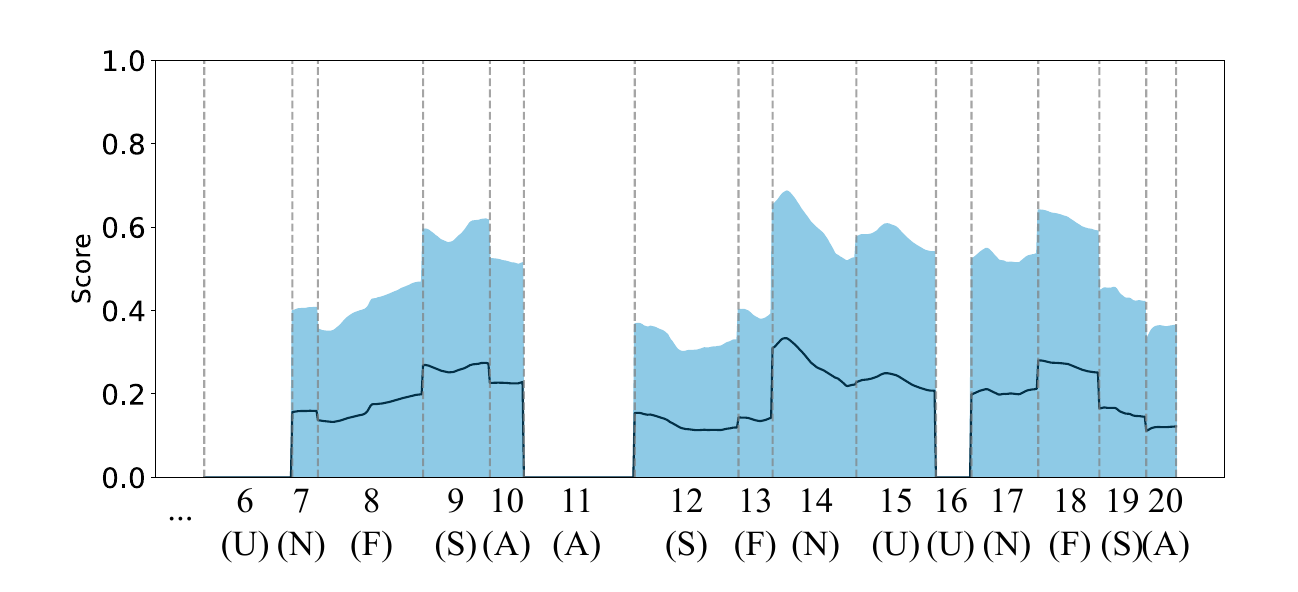}%
           }
            % \vspace{0.5cm}

           \subfloat[Session 3]{%
              \includegraphics[width=\linewidth]{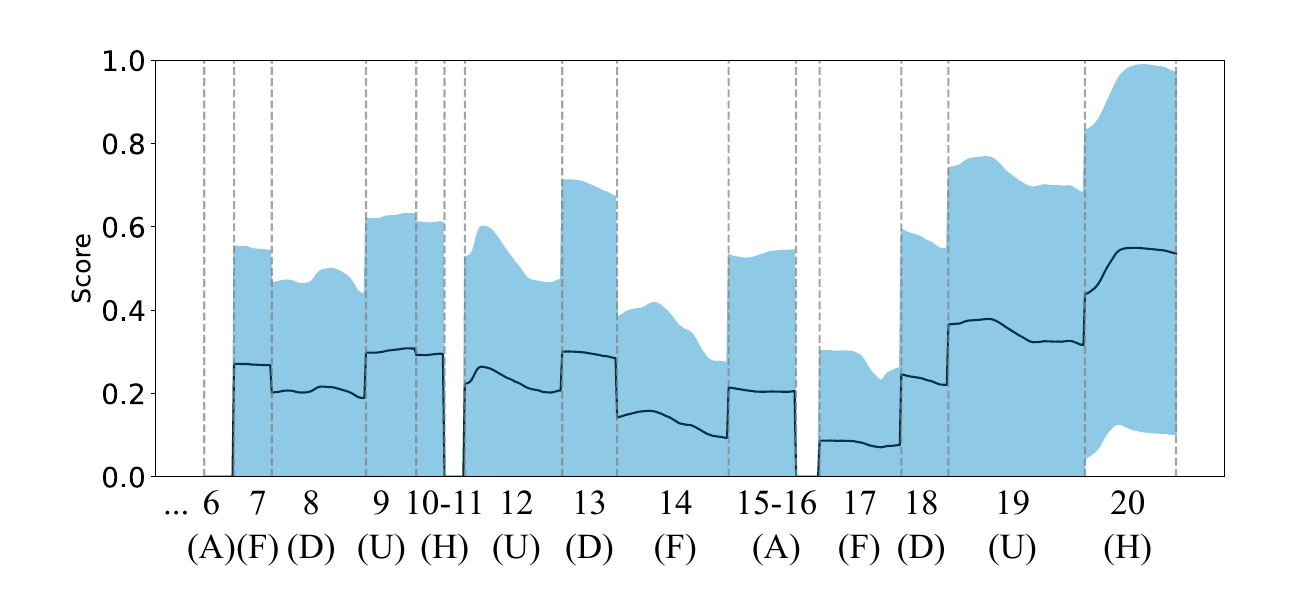}%
           }
           % \hspace{0.2cm} 
           
           \subfloat[Session 4]{%
              \includegraphics[width=\linewidth]{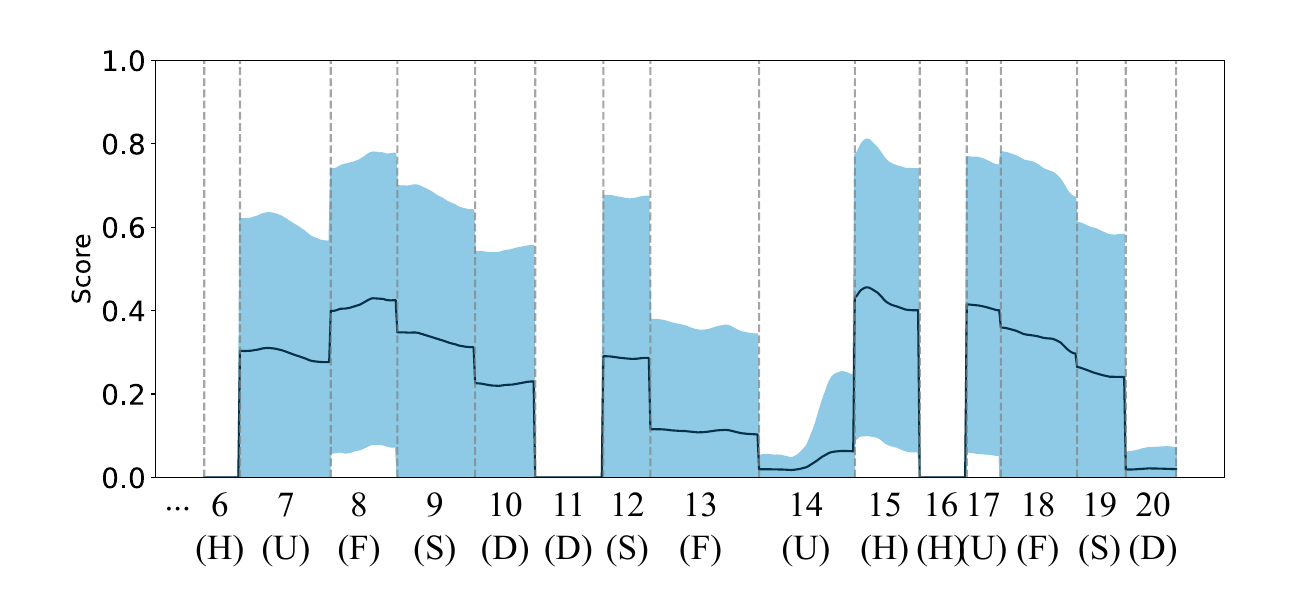}%
           }
           \caption{Averaged priming score for SEED-VII dataset. The priming score for each session is calculated independently as they do not follow the same sequence of emotions.}
           \label{fig:seed7-average-score}          
\end{figure} 

Affective priming has a different impact depending on the sequential order of emotional stimuli. For this reason, we visualize the average APS value for sequences that follow the same order of emotional stimuli. We applied the APS method to all sequences in the dataset and then averaged the obtained values across sequences. For example, assuming a sequence has $3000$ data points, we calculate the corresponding APS values for each data point and then average these values across the other sequences in the dataset.

In SEED, the sequence of emotions remains consistent across the three recorded sessions. Therefore, we compute the average affective priming scores across all sessions and participants. The results are presented in Fig.~\ref{fig:seed-average-score}.  
In SEED-VII, the sequential order of induced emotions varies across sessions. Therefore, the results for this benchmark are reported separately for each session.  
As done with SEED, we averaged the computed affective priming score across all participants in each session. The results are illustrated in Fig.~\ref{fig:seed7-average-score}.  

We highlight the variability in trial duration within the sequence, represented by the width between the dashed horizontal lines in Fig.~\ref{fig:seed-average-score} and Fig.~\ref{fig:seed7-average-score}. In the SEED dataset, this width remains mostly constant, indicating that approximately the same number of data points is available for each trial in the sequence. In contrast, in SEED-VII, the width between trials is more irregular, with some trials having fewer data points. This difference arises from the varying duration of the stimuli used in each dataset. In SEED, each stimulus lasts approximately four minutes on average, whereas in SEED-VII, the durations range from two to four minutes. This variability is reflected in the results presented in the next section.
 
\subsection{Training models on priming-free data sequences}

\begin{table}[]
    \caption{Configuration for the models used in this study.} \label{tab:config}
    \begin{tabular}{@{}cc@{}}
    \toprule
    Model & Configuration \\ \midrule
    LSTM & hidden\_dim = 64 \hspace{0.2cm} n\_layers = 1 \hspace{0.2cm} dropout = 0.3 \\
    Transformer & d\_model = 128 \hspace{0.2cm} n\_heads = 1 \hspace{0.2cm} n\_layers = 1 \\ \bottomrule
    \end{tabular}
\end{table}

The second part evaluates the effectiveness of the proposed method in detecting data points affected by affective priming. To do so, we train models on the original sequences and compare their performance with the same sequences after filtering out data points whose APS values exceed a predefined threshold. A core contribution of this work is the introduction of a metric to quantify the priming effect. This metric, named \textit{Priming Error}~(PE), is defined as follows:

% \begin{equation}
%     \label{eq:priming_error}
%     \resizebox{\columnwidth}{!}{$
%     PE = \frac{1}{N} \sum_{i=1}^{N} \delta_i, \text{~where } \delta_i = 
%     \begin{cases} 
%         1 & \text{if } \hat{y}_{i} = y_{i-1, k} \; \& \; y_{i-1, k} \neq y_{i,k} \\
%         0 & \text{otherwise}
%     \end{cases}
%     $}
% \end{equation}

% Hadi:
% \begin{equation}\small
%     \label{eq:priming_error}
%     \resizebox{\columnwidth}{!}{$
%     PE = \frac{1}{N} \sum_{i=1}^{N} \delta_i, \text{~~~~where } \delta_i = 
%     \begin{cases}
%         1 & \text{~if } \hat{y}_{i} = y_{i-1, k} \; \\ 
%         ~~~&  ~~\& \; y_{i-1, k} \neq y_{i,k} \\
%         0 & \text{otherwise}
%     \end{cases}
%     $}
% \end{equation}

\begin{equation} %\small
    \label{eq:priming_error}
    \resizebox{\columnwidth}{!}{$
    PE = \frac{1}{N} \sum_{i=1}^{N} \delta_i, \text{~where } \delta_i = 
    \begin{cases}
        1 & \text{if } \hat{y}_{i} = y_{i-1, k} \; \\ 
        ~~~&  ~~~\& \; y_{i-1, k} \neq y_{i,k} \\
        0 & \text{otherwise}
    \end{cases}
    $}
\end{equation}

The objective of Eq.~\ref{eq:priming_error} is to count the number of misclassifications in which the predicted label of a trial incorrectly matches the class of the previous trial in the sequence. For example, if a test data point labeled as \textit{neutral} is misclassified as \textit{sadness}, the class corresponding to the previous trial, this data point contributes to the PE metric. This metric serves as an indicator that the proposed data-driven method can effectively produce priming-free data sequences. Intuitively, lower PE values on the test set suggest that the model was trained on data less affected by the priming effect, thereby validating the proposed method.

To ensure the robustness of our method to identify priming-affected data points, we train two different sequential models. Specifically, we employ two well-known architectures: LSTM and Transformer (see Table~\ref{tab:config}). Given the limited number of available data points per sequence, we adjust the model configurations accordingly. We use the precomputed DE features provided in the benchmarks to construct sequences as model inputs~(see section~\ref{sec:datasets}). Each input sample is formed by concatenating consecutive DE feature vectors, enabling the model to capture temporal patterns and transitions between trials within a sequence. The models are trained with an initial learning rate of $1\mathrm{e}{-4}$ and a weight decay of $1\mathrm{e}{-5}$ for regularization. The batch size is set to $8$. For our experiments, we train a model for each subject-session combination, with each combination trained $5$ times using different random initializations (\textit{seeds}) to ensure statistical robustness. To promote reproducibility, we will make our code publicly available.

Based on findings in the literature~\cite{picard2001toward}, physiological signals vary across days and subjects. To prevent such variability from affecting the interpretation of the results, we train an independent model for each sequence in the dataset. In each dataset, a sequence is defined as the recording of a subject during one session (see Section~\ref{sec:datasets}).

Each model is trained on the first $9$ trials in SEED and the first $15$ trials in SEED-VII. For each trained model, the priming error~(PE) is evaluated on the test set, i.e., the remaining trials in the sequence. In SEED, results are aggregated across all sessions, as all recorded sessions follow the same sequence of emotional trials. In SEED-VII, results are reported separately for each session since each follows a distinct sequence of emotional trials. 

As previously mentioned, we compare model performance on the original sequences and their priming-free counterparts. To remove data points affected by the priming effect, we apply a threshold of $0.7$ on the APS values. This threshold balances the trade-off between retaining enough data to train the model and filtering out the most strongly affected data points.
The SEED dataset provides approximately $3300$ data points per sequence, whereas the SEED-VII dataset contains around $900$ data points per sequence. 
When applying this threshold, the average removed percentage of data points is $9\%$ and $10\%$ for the SEED and SEED-VII dataset, respectively.

% \begin{table}[t]
% \caption{Average priming error~(PE) on the test set~(trials 10 to 15) obtained in SEED. ($^{*}$represents statistical significance)}\label{tab:pe-seed}
% \centering
% \begin{tabular}{@{}lll@{}}
% \toprule
%  & Original & Priming-free \\ \midrule
% LSTM & $17.29\pm11.68$ & $\mathbf{8.41\pm6.63}^{*}$ \\
% Transformer & $14.20\pm10.01$ & $\mathbf{8.66\pm6.77}^{*}$ \\ \bottomrule
% \end{tabular}
% \end{table}

\begin{table}[t]
\caption{Average priming error~(PE) on the test set~(trials 10 to 15) obtained in SEED. ($^{*}$represents statistical significance)}\label{tab:pe-seed}
\centering
\begin{tabular}{@{}lll@{}}
\toprule
 & Original & Priming-free \\ \midrule
LSTM & \begin{tabular}[c]{@{}c@{}} $17.29\pm11.68$ \\ ($77.47\pm12.68$)\end{tabular} & \begin{tabular}[c]{@{}c@{}}$\mathbf{8.41\pm6.63}^{*}$ \\ ($83.67\pm09.48$)\end{tabular} \\ [0.5cm]
Transformer & \begin{tabular}[c]{@{}c@{}}$14.20\pm10.01$ \\ ($80.72\pm11.24$)\end{tabular} & \begin{tabular}[c]{@{}c@{}}$\mathbf{8.66\pm6.77}^{*}$ \\ ($84.71\pm09.21$)\end{tabular} \\ \bottomrule
\end{tabular}
\end{table}

% \begin{table}[t]
% \caption{Average priming error~(PE) on the test set~(trials 16 to 20) obtained in SEED-VII. (Sessions 1 and 2)}\label{tab:pe-seed7-sess12}
% \centering
% \resizebox{\columnwidth}{!}{  % Resize table to fit one column
% \begin{tabular}{@{}cccccc@{}}
% \toprule
%  & \multicolumn{2}{c}{Session 1} & \multicolumn{2}{c}{Session 2} \\ \midrule
%  & Original & Priming-free & Original & Priming-free \\
% LSTM & $13.61\pm7.78$ & $\mathbf{11.85\pm9.11}$ & $12.67\pm14.03$ & $\mathbf{9.58\pm12.36}$ \\
% Transformer & $12.96\pm8.94$ & $\mathbf{12.52\pm9.92}$ & $15.70\pm16.78$ & $\mathbf{13.48\pm13.14}$ \\ \bottomrule
% \end{tabular}
% }
% \end{table}

\begin{table}[t]
\caption{Average priming error~(PE) on the test set~(trials 16 to 20) obtained in SEED-VII. (Sessions 1 and 2)}\label{tab:pe-seed7-sess12}
\centering
\resizebox{\columnwidth}{!}{  % Resize table to fit one column
\begin{tabular}{@{}cccccc@{}}
\toprule
 & \multicolumn{2}{c}{Session 1} & \multicolumn{2}{c}{Session 2} \\ \midrule
 & Original & Priming-free & Original & Priming-free \\
LSTM & \begin{tabular}[c]{@{}c@{}}$13.61\pm7.78$ \\ ($33.12\pm13.19$)\end{tabular} & \begin{tabular}[c]{@{}c@{}} $\mathbf{11.85\pm9.11}$ \\ ($32.80\pm15.84$)\end{tabular} & \begin{tabular}[c]{@{}c@{}} $12.67\pm14.03$ \\ ($40.63\pm18.52$)\end{tabular} & \begin{tabular}[c]{@{}c@{}} $\mathbf{9.58\pm12.36}$ \\ ($43.05\pm18.74$)\end{tabular} \\ [0.5cm]
Transformer & \begin{tabular}[c]{@{}c@{}} $12.96\pm8.94$ \\ ($37.23\pm15.56$)\end{tabular} & \begin{tabular}[c]{@{}c@{}} $\mathbf{12.52\pm9.92}$ \\ ($34.96\pm15.73$)\end{tabular} & \begin{tabular}[c]{@{}c@{}} $15.70\pm16.78$ \\ ($48.90\pm15.68$)\end{tabular} & \begin{tabular}[c]{@{}c@{}} $\mathbf{13.48\pm13.14}$ \\ ($49.48\pm19.14$)\end{tabular} \\ \bottomrule
\end{tabular}
}
\end{table}

% \begin{table}[t]
% \caption{Average priming error~(PE) on the test set~(trials 16 to 20) obtained in SEED-VII. (Sessions 3 and 4)~($^{*}$represents statistical significance)}\label{tab:pe-seed7-sess34}
% \centering
% \resizebox{\columnwidth}{!}{  % Resize table to fit one column
% \begin{tabular}{@{}cccccc@{}}
% \toprule
%  & \multicolumn{2}{c}{Session 3} & \multicolumn{2}{c}{Session 4} \\ \midrule
%  & Original & Priming-free & Original & Priming-free \\
% LSTM & $19.29\pm16.94$ & $\mathbf{8.33\pm9.74^{*}}$ & $13.01\pm10.73$ & $\mathbf{8.32\pm10.30}$ \\
% Transformer & $15.45\pm12.00$ & $\mathbf{8.27\pm11.14}^{*}$ & $9.98\pm11.09$ & $\mathbf{9.11\pm10.23}$ \\ \bottomrule
% \end{tabular}
% }
% \end{table}
\begin{table}[t]
\caption{Average priming error~(PE) on the test set~(trials 16 to 20) obtained in SEED-VII. (Sessions 3 and 4)~($^{*}$represents statistical significance)}\label{tab:pe-seed7-sess34}
\centering
\resizebox{\columnwidth}{!}{  % Resize table to fit one column
\begin{tabular}{@{}cccccc@{}}
\toprule
 & \multicolumn{2}{c}{Session 3} & \multicolumn{2}{c}{Session 4} \\ \midrule
 & Original & Priming-free & Original & Priming-free \\
LSTM & \begin{tabular}[c]{@{}c@{}} $19.29\pm16.94$ \\ ($47.60\pm22.89$)\end{tabular} & \begin{tabular}[c]{@{}c@{}} $\mathbf{8.33\pm9.74^{*}}$ \\ ($53.11\pm27.08$)\end{tabular} & \begin{tabular}[c]{@{}c@{}} $13.01\pm10.73$ \\ ($44.44\pm22.46$)\end{tabular} & \begin{tabular}[c]{@{}c@{}} $\mathbf{8.32\pm10.30}$ \\ ($40.10\pm19.89$)\end{tabular} \\ [0.5cm]
Transformer & \begin{tabular}[c]{@{}c@{}} $15.45\pm12.00$ \\ ($51.27\pm16.60$)\end{tabular} & \begin{tabular}[c]{@{}c@{}} $\mathbf{8.27\pm11.14}^{*}$ \\ ($57.50\pm21.25$)\end{tabular} & \begin{tabular}[c]{@{}c@{}} $9.98\pm11.09$ \\ ($45.94\pm16.99$)\end{tabular} & \begin{tabular}[c]{@{}c@{}} $\mathbf{9.11\pm10.23}$ \\ ($38.91\pm19.45$)\end{tabular} \\ \bottomrule
\end{tabular}
}
\end{table}

In Table~\ref{tab:pe-seed}, we present the priming error~(PE) for the test trials in the SEED dataset. The columns \textit{Original} and \textit{Priming-free} report the average PE across the test trials, where the model was trained using either the original or the priming-free data sequence, respectively. We report similar results for the SEED-VII dataset in tables~\ref{tab:pe-seed7-sess12}-\ref{tab:pe-seed7-sess34}. Since in the SEED-VII dataset, sessions follow different emotional orders for the induced stimuli, we report results for each session independently. Additionally, we provide the corresponding average accuracy for each case.

The results reported in both datasets show a consistent reduction of the priming error when models are trained on priming-free sequences. Thus, these results confirm the efficiency of the APS method to detect data points affected by the priming effect.

%%%%%%%%%%
% Table~\ref{tab:pe-seed} presents the priming error obtained for the models trained on the SEED dataset. The columns \textit{Original} and \textit{Priming-free} show the average priming error~(PE) across all trained models, where the model is trained using either the original or the priming-free data sequence, respectively.  

% In the \textit{priming-free} column, the results correspond to the case where a threshold value of $0.7$ is applied, meaning that all samples with an APS value higher than $0.7$ are removed from the sequence before model training and testing. This threshold was chosen to balance the trade-off between removing primed samples and retaining sufficient data for training. The SEED dataset provides approximately $3300$ samples per sequence, whereas the SEED-VII dataset contains around $900$ samples per sequence.  

% Tables~\ref{tab:pe-seed7-sess12} and~\ref{tab:pe-seed7-sess34} present the results obtained for the four different sessions in SEED-VII. In all tables, we highlight the result with the lowest average PE across all trained models.

\begin{figure}
    \centering
    \includegraphics[width=\linewidth]{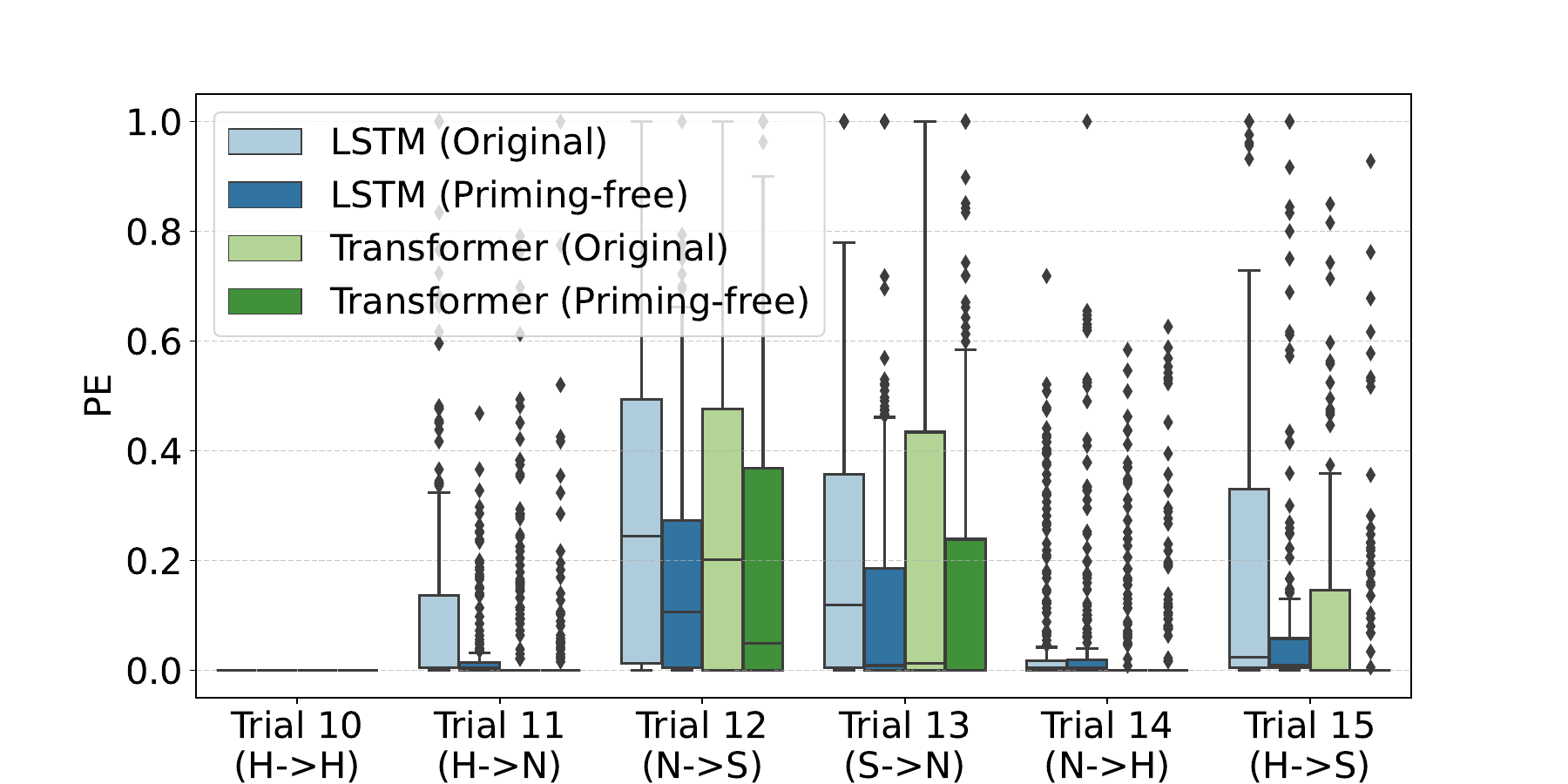}
    \caption{Distribution of the priming error (PE) for each trial in the test sequence in the SEED dataset.}
    \label{fig:boxplot-seed}
\end{figure}

\begin{figure*}[t]
        \centering
           \subfloat[Session 1]{%
              \includegraphics[width=0.49\linewidth]{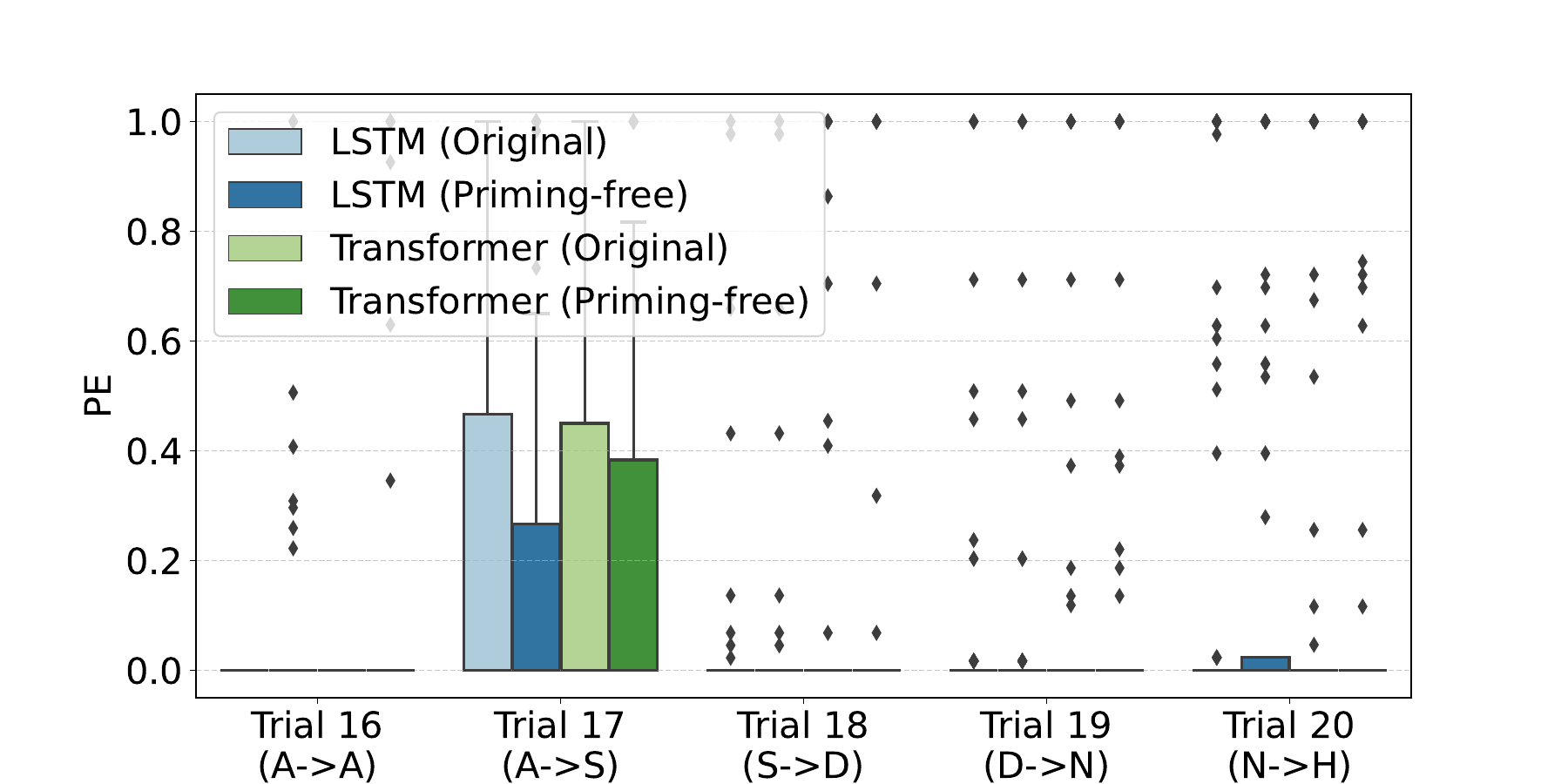}%
           } 
           % \hspace{0.2cm}
           \subfloat[Session 2]{%
              \includegraphics[width=0.49\linewidth]{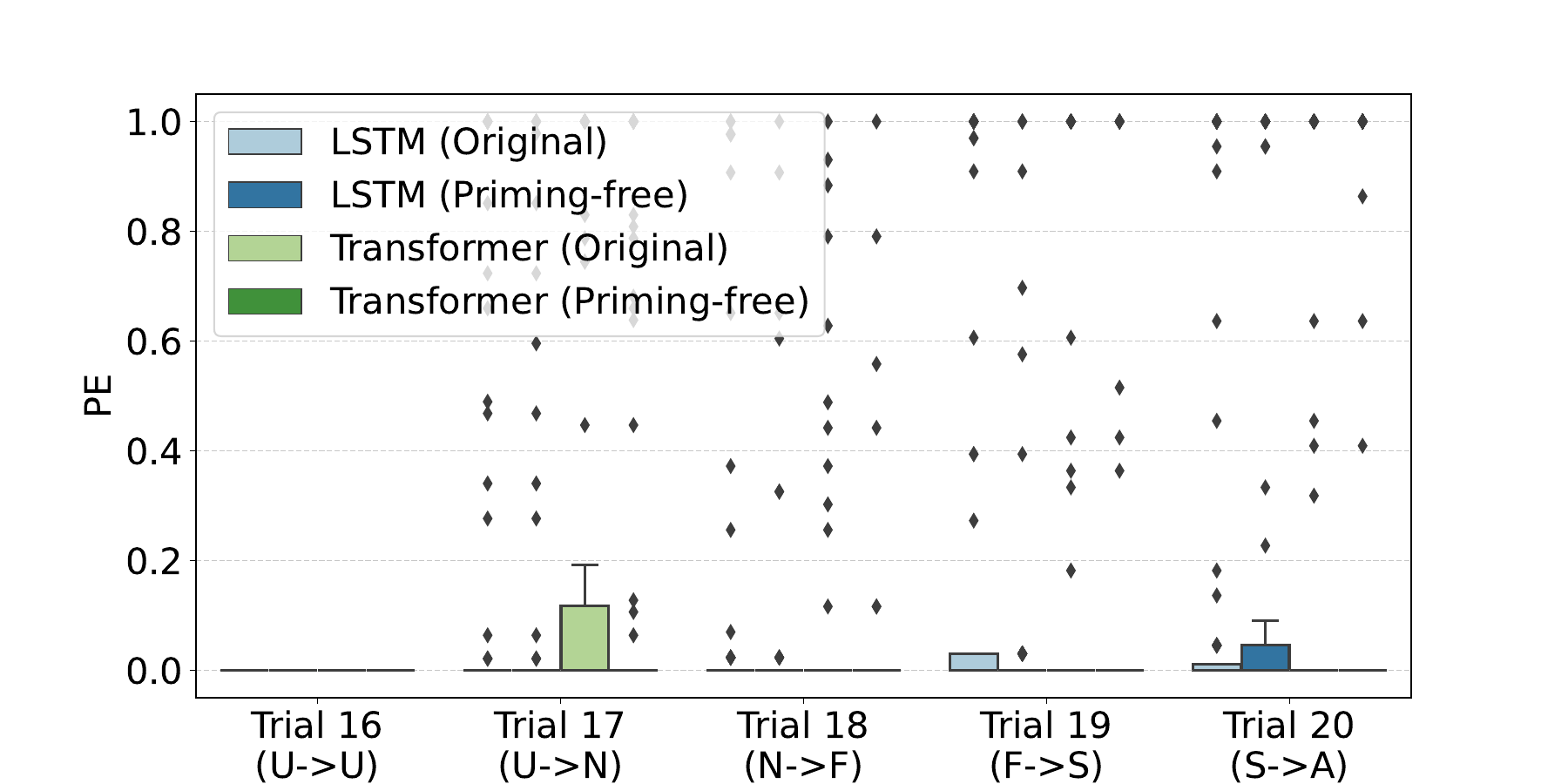}%
           }
        
            \vspace{0.5cm}
        
           \subfloat[Session 3]{%
              \includegraphics[width=0.48\linewidth]{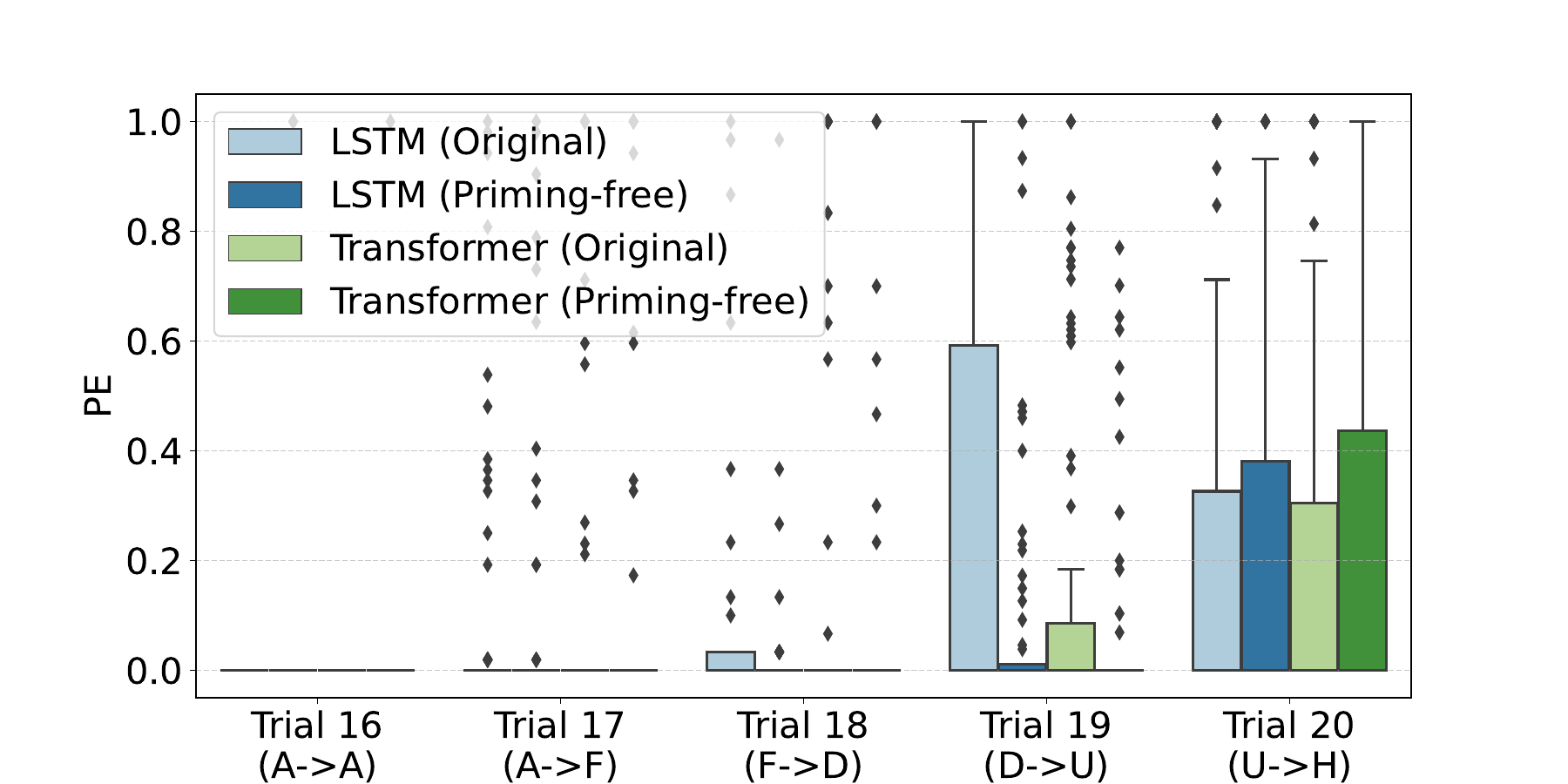}%
           }
           \hspace{0.2cm} 
           \subfloat[Session 4]{%
              \includegraphics[width=0.48\linewidth]{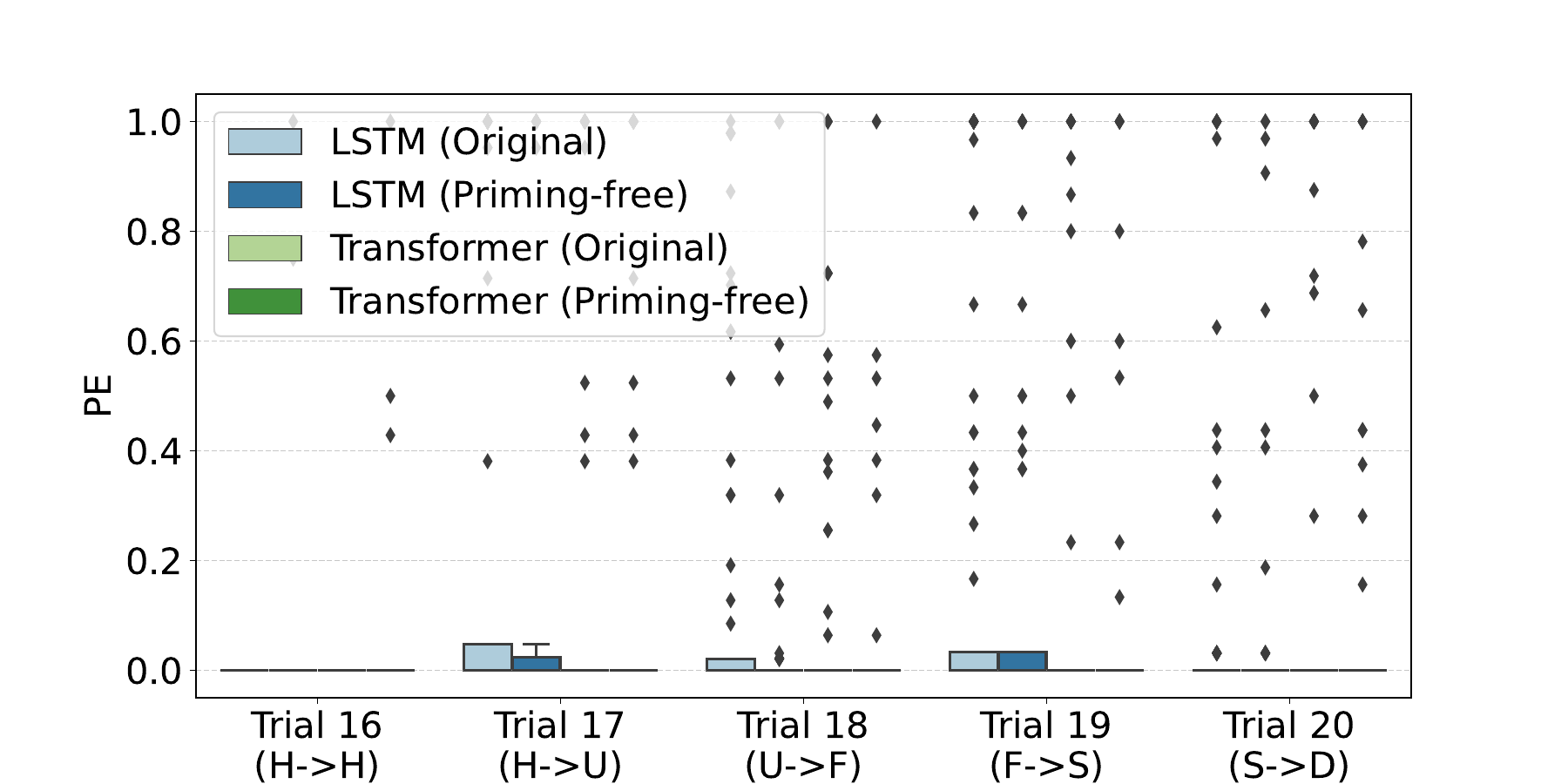}%
           }
           \caption{Distribution of the priming error (PE) for each trial in the test sequence in the SEED-VII dataset.}
           \label{fig:boxplot-seed7}
\end{figure*} 

We further investigate the effect of removing priming-affected data points by disaggregating the previously reported results. In this analysis, we aim to illustrate the distribution of the PE metric across individual trials in the test sequence. To achieve this, we compute the PE value for each trial in every trained model.  
The resulting distributions are visualized using box plots, as shown in Fig.~\ref{fig:boxplot-seed} for the SEED dataset and Fig.~\ref{fig:boxplot-seed7} for SEED-VII. The variation in PE values is notably heterogeneous, as each sequence in the dataset, i.e., each subject in each recorded session, is affected by priming to a different extent. Interestingly, the trials with higher average PE values correspond to those where the proposed method also reports higher average APS values in the sequence. For example, in the SEED dataset, trial 12 shows higher PE values compared to other trials in the test sequence, and Fig.~\ref{fig:seed-average-score} shows a higher APS value for trial 12 as well. A similar pattern is observed in the SEED-VII dataset, specifically in trial 17 of session 1 and trial 20 of session 3. These results highlight the effectiveness of the proposed method in identifying priming-affected data points in sequential datasets.

\section{Discussion}

% Discussion points: 
% 1. Data-centric affective priming quantification
% 2. System limitations
% 3. Future work

\textbf{Data-centric affective priming quantification.} To the best of our knowledge, no previous work has addressed the challenge of affective priming from a data-centric perspective. This work highlights its importance and proposes a method that not only prevents performance degradation in models but also enables visualization of the priming effect in data sequences. This visualization can serve as a tool to enhance the design of dataset collection protocols.

\textbf{System limitations.} Although the results show a decrease in Priming Error (PE) across all models for both the SEED and SEED-VII datasets, the difference in accuracy between these databases is noteworthy. In SEED, both metrics (PE and accuracy) improve, while in SEED-VII, some sessions (e.g., sessions 1 and 4) show worse accuracy on priming-free sequences. We attribute this to differences in the number of available data points per sequence in each dataset. Consequently, the effectiveness of the proposed method is limited by the amount of available data.

\textbf{Future work.} To address ambiguity in systems of this kind, it is essential to incorporate all relevant information during the modeling phase. While this study introduces a method to quantify the priming effect in the data, incorporating such information during model training would be necessary to address the issue. Future work will explore strategies to effectively integrate such information.

\section{Conclusion}

This study explored the problem of affective priming from a data-centric perspective and introduced a method to detect data points influenced by this effect in sequential datasets.
We propose the Affective Priming Score (APS), a data-driven method to identify data points that are more likely to be impacted by priming. To evaluate the effectiveness of APS in detecting priming in data points, we train models using two data types: the original sequence and its priming-free version. Given the absence of ground truth for affective priming, we introduce the priming effect metric, which quantifies the proportion of misclassified data points that are incorrectly predicted as the emotion from the preceding trial in the sequence. Experimental results demonstrate the effectiveness of APS, as it significantly reduces this metric.

Beyond its application in cleaning data sequences, this tool has the potential to improve the design of data collection protocols. By analyzing how different sequences of emotional stimuli influence collected data, researchers can optimize experimental setups to minimize priming effects.

Addressing label ambiguity remains a critical challenge in affective computing. This work contributes to ongoing efforts in the field by providing a systematic method for detecting and mitigating affective priming at the data level, thereby improving the reliability of emotion recognition models and the overall quality of affective computing datasets.

%%  ---- START: Added 2022-03-15 for ACII2022 onwards -----
\section*{Ethical Impact Statement}

% The data that have been used in this work comes from two public benchmarks available on the internet.
% The data provided in these benchmarks corresponds to EEG information, which makes the subjects unidentifiable. 
% Additionally, the authors reflect on the potential negative societal impact.
% There is no direct negative impact with the solution provided in this work, however, malicious use of it could address the privacy of the monitored users.
% The data used in this work may present a cultural bias, which may limit the generalization capabilities of the presented models. 
% However, the scope of this study was limited to subject-dependent models. 

The data used in this work originate from two publicly available benchmark datasets found online. 
These datasets contain EEG recordings, which do not allow for the identification of individual subjects.
The authors have also considered the potential for negative societal impact. 
While the proposed solution does not pose any immediate risk, it could potentially be misused in ways that compromise the privacy of monitored individuals. 
Additionally, the datasets may contain cultural biases, which could affect the generalizability of the developed models. 
Nonetheless, the focus of this study was limited to subject-dependent modeling.

% At ACII2023, we are requiring a {\color{red} mandatory} field to be filled in on the manuscript submission portal to discuss any issues regarding the Ethical Impact of their work. This allows authors to reflect on the ethical impact of the work, which could include societal and environmental impact.

% We also {\color{red}highly recommend} authors to include a corresponding Ethical Impact Statement in their papers to discuss these issues. If authors choose to do so, this section should be un-numbered (like the Acknowledgement), and come after the conclusions, but before the Acknowledgement and References. (For example, the text in this section can be used to answer the mandatory field on the manuscript submission portal.)

% Please see the Conference Submission Guidelines on the conference website for up-to-date information and guidelines.

%%  ---- END: Added 2022-03-15 for ACII2022 onwards -----

\bibliographystyle{ieeetr}
%\bibliography{ref}

\begin{thebibliography}{10}

\bibitem{bargh2000mind}
J.~A. Bargh and T.~L. Chartrand, ``The mind in the middle,'' {\em Handbook of
  research methods in social and personality psychology}, vol.~2, pp.~253--285,
  2000.

\bibitem{shen2019unintentional}
J.~H. Shen, A.~Lapedriza, and R.~W. Picard, ``Unintentional affective priming
  during labeling may bias labels,'' in {\em 2019 8th International Conference
  on Affective Computing and Intelligent Interaction (ACII)}, pp.~587--593,
  IEEE, 2019.

\bibitem{martinez2023analyzing}
L.~Martinez-Lucas, A.~Salman, S.-G. Leem, S.~G. Upadhyay, C.-C. Lee, and
  C.~Busso, ``Analyzing the effect of affective priming on emotional
  annotations,'' in {\em 2023 11th International Conference on Affective
  Computing and Intelligent Interaction (ACII)}, pp.~1--8, IEEE, 2023.

\bibitem{maestro2023stress}
E.~G. Maestro, H.~Banaee, and A.~Loutfi, ``Stress lingers: Recognizing the
  impact of task order on design of stress and emotion detection systems,'' in
  {\em 2023 IEEE EMBS Special Topic Conference on Data Science and Engineering
  in Healthcare, Medicine and Biology}, pp.~175--176, IEEE, 2023.

\bibitem{hochreiter1996lstm}
S.~Hochreiter and J.~Schmidhuber, ``Lstm can solve hard long time lag
  problems,'' {\em Advances in neural information processing systems}, vol.~9,
  1996.

\bibitem{vaswani2017attention}
A.~Vaswani, N.~Shazeer, N.~Parmar, J.~Uszkoreit, L.~Jones, A.~N. Gomez,
  {\L}.~Kaiser, and I.~Polosukhin, ``Attention is all you need,'' {\em Advances
  in neural information processing systems}, vol.~30, 2017.

\bibitem{du2020efficient}
X.~Du, C.~Ma, G.~Zhang, J.~Li, Y.-K. Lai, G.~Zhao, X.~Deng, Y.-J. Liu, and
  H.~Wang, ``An efficient lstm network for emotion recognition from
  multichannel eeg signals,'' {\em IEEE Transactions on Affective Computing},
  vol.~13, no.~3, pp.~1528--1540, 2020.

\bibitem{sakalle2021lstm}
A.~Sakalle, P.~Tomar, H.~Bhardwaj, D.~Acharya, and A.~Bhardwaj, ``A lstm based
  deep learning network for recognizing emotions using wireless brainwave
  driven system,'' {\em Expert Systems with Applications}, vol.~173, p.~114516,
  2021.

\bibitem{meng2023eeg}
M.~Meng, Y.~Zhang, Y.~Ma, Y.~Gao, and W.~Kong, ``Eeg-based emotion recognition
  with cascaded convolutional recurrent neural networks,'' {\em Pattern
  Analysis and Applications}, vol.~26, no.~2, pp.~783--795, 2023.

\bibitem{ouyang2025daeegvit}
Y.~Ouyang, Y.~Liu, L.~Shan, Z.~Jia, D.~Qian, T.~Zeng, and H.~Zeng, ``Daeegvit:
  A domain adaptive vision transformer framework for eeg cognitive state
  identification,'' {\em Biomedical Signal Processing and Control}, vol.~100,
  p.~107019, 2025.

\bibitem{zhang2024transformer}
X.~Zhang and X.~Cheng, ``A transformer convolutional network with the method of
  image segmentation for eeg-based emotion recognition,'' {\em IEEE Signal
  Processing Letters}, vol.~31, pp.~401--405, 2024.

\bibitem{song23Trans}
Y.~Song, Q.~Zheng, B.~Liu, and X.~Gao, ``Eeg conformer: Convolutional
  transformer for eeg decoding and visualization,'' {\em IEEE Transactions on
  Neural Systems and Rehabilitation Engineering}, vol.~31, 2023.

\bibitem{seed}
W.-L. Zheng and B.-L. Lu, ``Investigating critical frequency bands and channels
  for eeg-based emotion recognition with deep neural networks,'' {\em IEEE
  Transactions on autonomous mental development}, vol.~7, no.~3, pp.~162--175,
  2015.

\bibitem{seed7}
W.-B. Jiang, X.-H. Liu, W.-L. Zheng, and B.-L. Lu, ``Seed-vii: A multimodal
  dataset of six basic emotions with continuous labels for emotion
  recognition,'' {\em IEEE Transactions on Affective Computing}, 2024.

\bibitem{deap}
S.~Koelstra, C.~Muhl, M.~Soleymani, J.-S. Lee, A.~Yazdani, T.~Ebrahimi, T.~Pun,
  A.~Nijholt, and I.~Patras, ``Deap: A database for emotion analysis; using
  physiological signals,'' {\em IEEE transactions on affective computing},
  vol.~3, no.~1, pp.~18--31, 2011.

\bibitem{dreamer}
S.~Katsigiannis and N.~Ramzan, ``Dreamer: A database for emotion recognition
  through eeg and ecg signals from wireless low-cost off-the-shelf devices,''
  {\em IEEE journal of biomedical and health informatics}, vol.~22, no.~1,
  pp.~98--107, 2017.

\bibitem{picard2001toward}
R.~W. Picard, E.~Vyzas, and J.~Healey, ``Toward machine emotional intelligence:
  Analysis of affective physiological state,'' {\em IEEE transactions on
  pattern analysis and machine intelligence}, vol.~23, no.~10, pp.~1175--1191,
  2001.

\end{thebibliography}

\end{document}